\documentclass[final]{article}

\usepackage{aistats2020}
\usepackage{amsthm}
\usepackage[utf8]{inputenc} 
\usepackage[T1]{fontenc}    
\usepackage{hyperref}       
\usepackage{url}            
\usepackage{booktabs}       
\usepackage{amsfonts}       
\usepackage{nicefrac}       
\usepackage{microtype}      
\usepackage{lipsum}
\usepackage{amssymb,amsmath,amsfonts}
\usepackage{latexsym}
\usepackage{graphicx}
\usepackage{mathrsfs}
\usepackage{geometry}
\usepackage{fancyhdr}
\usepackage{booktabs}
\usepackage{amsmath}
\usepackage{amsfonts}
\usepackage{amssymb}
\usepackage{multirow}
\usepackage{array}
\usepackage{natbib}
\usepackage{bm}
\usepackage{graphics}

\def\x{\mathbf x}

\def\s{\mathbf s}

\def\W{\mathbf W}

\def\L{\mathcal{L}}

\def\V{\mathbb V}

\DeclareMathOperator{\sign}{sign} 
\DeclareMathOperator{\BN}{BN}
\DeclareMathOperator{\var}{\mathbb{V}}

\def \real{\rm I\!R}

\def \ind{\mathbf 1}
\def \L {\mathcal{L}}
\newtheorem{theorem}{\sffamily Theorem}

\usepackage[font={footnotesize,bf,sf}]{caption}
\usepackage{color}
\usepackage{listings}
\definecolor{gris}{rgb}{0.44,0.44,0.44}
\graphicspath{{Figures/}}
\PassOptionsToPackage{hyphens}{url}
\usepackage[utf8]{inputenc}
\usepackage{hyperref} 
\hypersetup{colorlinks,%
	citecolor={blue}, 
	urlcolor={blue},
	linkcolor={blue},
	breaklinks={true},
}

\begin{document}

\twocolumn[

\aistatstitle{How Does Batch Normalization Help Binary Training?}
\aistatsauthor{ Eyy\"ub Sari \And Mouloud Belbahri \And  Vahid Partovi Nia   }
\aistatsaddress{ Huawei Noah's Ark Lab \\ eyyub.sari@huawei.com \And  University of Montreal \\ 
Department of Math. Stats. \\ mouloud.belbahri@gmail.com \And Huawei Noah's Ark Lab \\ vahid.partovinia@huawei.com } ]

\begin{abstract}
Binary Neural Networks (BNNs) are difficult to train, and suffer from drop of accuracy. It appears in practice that BNNs fail to train in the absence of Batch Normalization (BatchNorm) layer. We find the main role of BatchNorm is to avoid exploding gradients in the case of BNNs. This finding suggests that the common initialization methods developed for full-precision networks are irrelevant to BNNs. We build a theoretical study on the role of BatchNorm in binary training, backed up by  numerical experiments.
\end{abstract}


\section{Introduction}

Deep Neural Networks (DNNs) are complex and resources-hungry, preventing them from being deployed on edge devices. Binary Neural Networks (BNNs) \citep{courbariaux2016bnn} are an attempt to alleviate these issues by adopting an extreme quantization scheme and constraint weight and activation to $\{-1,+1\}$. This extreme constraint often leads to under-fitting. Moreover, training networks with binary values does not match stochastic gradient descent mission, which is designed for continuous weight and activation. Consequently,  BNNs suffer from an accuracy drop compared to their full-precision counterpart.  Most of binary training in convolutional models include Batch Normalization (BatchNorm) layer \citep{ioffe2015batchnorm}. However, the BatchNorm layer is missing in most of recurrent networks, but \cite{ardakani2018learningrecbinter} report competitive quantized training by only adding BatchNorm to the gates. Here we argue why binary training without BatchNorm is infeasible.

 It is natural to take a step back and wonder how important BatchNorm component is. We show that Glorot initialization \citep{glorot2010init} that protects gradients from explosion or vanishing is useless for  BNNs, and this could be the reason why training BNNs using a pretrained model is preferred \citep{alizadeh2018empiricalbnn}. Briefly 
i) we show BNNs are  insensitive to Glorot's initialization parameters, so initializing weights from a distribution symmetric about zero and with an arbitrary variance is sufficient; ii) we demonstrate that BatchNorm  prevents exploding gradient, and this make the common full-precision initialization schemes pointless; iii) we break down BatchNorm components to centering and scaling, and show only mini-batch centering is required; the scale can be fixed to a certain constant. We only perform experiments on CIFAR-10, but experiments on other data sets lead to the same conclusion.


\section{Preliminaries}
 BNNs apply the $\sign$ function on full-precision values, only keeping the sign while discarding the magnitude. This allows up to $32\times$ memory-wise compression and yields fast inference via XNOR-Popcount operation compared to $\mathrm{float32}$-precision networks. 

Let $x \in \real$, the derivative of the $\sign$ function is required for backpropagation. An approximation, called the  \emph{clipped straight-through estimator} is 
$\frac{\partial \sign(x)}{\partial x} \approx \ind_{[-1,1]}(x),$ where $\ind(.)$ is the indicator function.

A random variable taking binary values $\{-1,1\}$ can be expressed as a transformed Bernoulli. The mathematical expectation and variance of this transformed Bernoulli variable $\tilde x$ is	
$$\mathbb{E}(\tilde x) =  2p - 1, \quad 
		\var(\tilde x) =  4p(1-p)
$$
As a consequence, the sum of such $\tilde x$'s is a  location-scale Binomial.  The binary random variable $\tilde x$ is symmetric about zero if $p={1\over 2}$. The sum of $K$ such binary variables is also symmetric about zero, but the variance evolves from $1$ to $K$. This property may require attention in initialization of binary networks, as the variance of such binary dot products need to be controlled.

It is well-known that BatchNorm facilitated  neural networks training. A common intuition suggests  BatchNorm  matches input and output distribution in their first and the second moments.  There are two other clues among others: \cite{ioffe2015batchnorm} claim that BatchNorm corrects covariate shift, and  \cite{santurkar2018bnoptim} show BatchNorm bounds the gradient and makes the optimization smoother. None of these arguments work for BNNs! The role of BatchNorm is to prevent exploding gradient.

Suppose 
a mini batch of size $B$ for a given neuron $k$. Let 
$\hat{\mu}_{k}, \hat{\sigma}_{k}$ be the mean and the standard deviation of the dot product, between inputs and weights, $s_{bk}, b=1,\dots B$. For a given layer $l$, BatchNorm is defined as $
			\BN(s_{bk}) \equiv z_{bk} = \gamma_k \hat{s}_{bk} + \beta_k,$
where $\hat{s}_{bk} = \frac{s_{bk} - \hat{\mu}_{k}}{\hat{\sigma}_{k}} $ is the standardized dot product and the pair $(\gamma_k$,  $\beta_k)$ is trainable, initialized with $(1, 0)$.

Given the objective function $\L(.)$, BatchNorm parameters are trained in backpropagation 
$$	    \frac{\partial \L}{\partial \beta_k} = \sum_{b=1}^{B}\frac{\partial \L}{\partial z_{bk}},
	    \quad
		\frac{\partial \L}{\partial \gamma_k} = \sum_{b=1}^{B} \frac{\partial \L}{\partial z_{bk}} \hat{s}_{bk},
$$
For a given layer $l$, it is easy to prove $\frac{\partial \L}{\partial s_{bk}}$ equals 
    \begin{equation}\label{eq:bnderiv}
		\frac{\gamma_k}{\hat{\sigma}_{k}} \Big(- \frac{1}{B}\sum_{b'=1}^{B} \frac{\partial \L}{\partial z_{b'k}} - \frac{\hat{s}_{bk}}{B}\sum_{b'=1}^{B} \frac{\partial \L}{\partial z_{b'k}} \hat{s}_{b'k} + \frac{\partial \L}{\partial z_{bk}} \Big).
    \end{equation}

We follow the same assumptions as in  \cite{glorot2010init}, i.e. weights and activations are  independent, and identically distributed (iid) and centred about zero. Formally, denote the dot product vector $\s_b^l \in {\real}^{K_l}$  of sample $b$ in layer $l$, with $K_l$ neurons. Let $f$ be the element-wise activation  function,  $\x_b$ be the input vector, $\W^l \in {\real}^{K_{l-1} \times K_l}$ with elements $\mathbf W^l = [w^l_{kk'}]$ be the weights matrix; one may use simply $w^l$ referring to an identically distributed elements of layer $l$. It is easy to verify 
\begin{eqnarray*}
			\frac{\partial \L}{\partial s^l_{bk}} &=& f'(s^l_{bk})\sum_{k'=1}^{K_{l+1}}w^{l+1}_{kk'}\frac{\partial \L}{\partial s^{l+1}_{bk'}}, \\
			\frac{\partial \L}{\partial w^l_{k'k}}&=&  \sum_{b=1}^Bs^{l-1}_{bk'}\frac{\partial \L}{\partial s^l_{bk}}.
\end{eqnarray*}

We assume that the feature element $x$ and the weight element $w $ are centred and iid, for a given neuron $k=1,\ldots, K_l$ at layer $l$.  We reserve $k$ to index the current neuron and use $k'$ for the previous or the next layer neuron depending on the context.  One can show 
				$\var(s^l_{bk}) = \var(x)\prod_{l'=1}^{l-1} K_{l'} \var(w^{l'}),$
where $\var(w^{l'})$ is the variance of the weight in layer $l'$. By applying similar mathematical mechanics, the variance of the gradient for a neuron is
$$			\var(\frac{\partial \L}{\partial s^l_{bk}}) = \var(\frac{\partial \L}{\partial s^L})\prod_{l'=l+1}^LK_{l'}\var(w^{l'}),
			$$
which explodes or vanishes depending on $\var(w^{l'})$. This is the main reason common full-precision initialization methods suggest $\var(w^l)= {1\over K_{l}}$.

\section{Effect of BatchNorm}
We first question the role of BatchNorm in BNNs, and demonstrate that BatchNorm has a different role compared to the one it has in full-precision models.  Most of BNNs contain BatchNorm layers, because  they are infeasible to train without BatchNorm. BNNs use the $\sign$ activation, therefore the scale of input distribution has no effect on the forward pass, and only the location plays a role. This observation is enlightening:  at inference time, a BatchNorm layer followed by $\sign$ is a threshold function. This motivates us to focus on centering more than scaling while initializing BNNs
\begin{equation}
\sign\{\mathrm{BN}(s_{bk})\} 
				= \begin{cases}
				+1& \text{if } s_{bk} \geq \lfloor \mu_{k} - \frac{\hat{\sigma}_{k}}{\gamma_k}\beta_k \rfloor, \\
				-1& \text{otherwise.}
				\end{cases}			
\label{eq:signBN}
\end{equation}
where $\lfloor . \rfloor$ is the floor to integer. 
From \eqref{eq:signBN} it is clear that controlling the variance has no fundamental effect on forward propagation if $s_{bk}$ is symmetric about zero. The term $b_k= \mu_{k} - \frac{\hat{\sigma}_{k}}{\gamma_k}\beta_k$ can be regarded as as a new trainable parameter, thus BatchNorm layer can be replaced by adding biases to the network to compensate. We are aware of dismissal of BatchNorm in forward pass but still BNNs do not converge in practice without BatchNorm. We suspect that BatchNorm layers help backpropagation and this is why \emph{only} BNNs with BatchNorm converge.

\section{BatchNorm and Initialization}
In presence of BatchNorm layers, one may show that weights variance has no impact on the forward pass as the output will be normalized after passing through BatchNorm. In another word, for a given layer $l$, 
$			 \var(\BN(s_{bk})) = \gamma_k^2.$
For any full-precision network,  BatchNorm affects backpropagation as 
		\begin{eqnarray}
		\var \Big(\frac{\partial \L}{\partial s^l_{bk}} \Big) 
			&=& \Big(\frac{\gamma^{l}_k}{B {\hat{\sigma}^l}_{k}} \Big)^2 \{B^2 + 2B - 1 + \var(\hat{s}_{bk}^{l^2}) \} \nonumber \\ 
			&&K_{l+1}\var(w^{l+1})\var \Big(\frac{\partial \L}{\partial s^{l+1}} \Big).
			\label{eq:backbnnvar}
		\end{eqnarray}		
Setting $\V(w^{l+1})={1\over K_{l+1}}$ still does not stabilize the variance of the gradient. Gradients are stabilized only if $\Big(\frac{\gamma^{l}_k}{B} \Big)^2 \{B^2 + 2B - 1 + \var(\hat{s}_{bk}^{l^2}) \} \approx 1$. Moving from full precision weight $w$ to binary weight $\tilde w = \sign (w)$ changes the situation dramatically. In other words initializing with a symmetric distribution about zero with any variance gives $\V(\tilde w^{l+1})=1,$ which does not cancel out the $K_{l+1}$ in \eqref{eq:backbnnvar}. Theorem~\ref{theo:batchnorm} shows BatchNorm approximately cancels $K_{l+1}$.
\begin{theorem} \label{theo:batchnorm}
Under common initialization assumptions, the gradient variance for BNNs without BatchNorm is
$$
		\var \Big(\frac{\partial \L}{\partial s^l_{bk}} \Big)  = \var\Big(\frac{\partial \L}{\partial s^L}\Big)\prod_{l'=l+1}^L K_{l'},$$
and with BatchNorm is  
$$ \var \Big(\frac{\partial \L}{\partial s^l_{bk}} \Big)  =\prod_{l'=l}^{L-1}\frac{K_{l'+1}}{K_{l'-1}}\var\Big(\frac{\partial \L}{\partial s^{L}}\Big)+o\left({1\over B^{1-\epsilon}} \right),
$$ for an arbitrary $0<\epsilon<1$.
\end{theorem}
\textbf{Proof of Theorem~1}. The proof for BNNs without BatchNorm is trivial, so we only provide the proof for BNNs \emph{with} BatchNorm. Here, $s^l_{bk}$ refers to a binary dot product and $\tilde w^l$ is a binary random variable representing the binary weight. We may suppose $\var(s^l_{bk}) \equiv {\hat{\sigma}^{l^2}}_{k} = K_{l-1}$, ${\hat{\mu}^l}_{k}=0$. One may prove $\var \Big(\frac{\partial \L}{\partial s^l_{bk}} \Big) 
			= \Big(\frac{\gamma^{l}_k}{B {\hat{\sigma}^l}_{k}} \Big)^2 \{B^2 + 2B - 1 + \var(\hat{s}_{bk}^{l^2}) \} K_{l+1}\var(\tilde w^{l+1})\var \Big(\frac{\partial \L}{\partial s^{l+1}} \Big) $. It is also possible to verify  $\var(\hat{s}_{bk}^{l^2})$ is bounded by $K_{l-1}-1$; the proofs appear in  Appendix. Suppose  $\gamma^{l}_k=1$ at initialization so
\begin{eqnarray*}
    &&\Big(\frac{\gamma^{l}_k}{B {\hat{\sigma}^l}_{k}} \Big)^2 \{B^2 + 2B - 1 + \var(\hat{s}_{bk}^{l^2}) \}\\ &=& \Big(\frac{1}{B \sqrt{K_{l-1}}} \Big)^2 \{B^2 + 2B - 1 + \var(\hat{s}_{bk}^{l^2}) \}\\
    &=&  \frac{\{B^2 + 2B - 1 + \var(\hat{s}_{bk}^{l^2}) \}}{K_{l-1}B^2}\\ &=&  \frac{1}{K_{l-1}} \left\{1 + o\left({1\over B^{1-\epsilon}}\right)\right\}.
\end{eqnarray*}
Therefore, 
\begin{eqnarray*}
    && \var \Big(\frac{\partial \L}{\partial s^l_{bk}} \Big) \\
			&=& \Big(\frac{\gamma^{l}_k}{B {\hat{\sigma}^l}_{k}} \Big)^2 \{B^2 + 2B - 1 + \var(\hat{s}_{bk}^{l^2}) \} \\ &&~~ K_{l+1}\var(\tilde w^{l+1})\var \Big(\frac{\partial \L}{\partial s^{l+1}} \Big)  \\ 
			\end{eqnarray*}
			After slight re-arrangement
			\begin{eqnarray*}
			&=& \frac{K_{l+1}}{K_{l-1}}\left\{1 + o\left({1\over B^{1-\epsilon}}\right)\right\}\var\Big(\frac{\partial \L}{\partial s^{l+1}} \Big)\\
			&=& \var\Big(\frac{\partial \L}{\partial s^{L}}\Big)\prod_{l'=l}^{L-1}\frac{K_{l'+1}}{K_{l'-1}}+o\left({1\over B^{1-\epsilon}} \right).
\end{eqnarray*}

There are two implications of Theorem~\ref{theo:batchnorm}. i)  BatchNorm corrects exploding gradients in BNNs as the layer width ratio  ${K_{l+1}\over K_{l-1}}\approx 1$ in common neural models.  If this ratio diverges from unity binary training is problematic even with  BatchNorm. ii) One may initialize the weights of a BNN from any distribution symmetric and centred about  zero. We also demonstrate this analytical results are aligned with numerical experiments.

\section{Experiments} \label{experiments}
We verify our analytical finding on the CIFAR-10 dataset \citep{krizhevsky2009cifar10} which consists of $50,000$ training images and $10,000$ test images. The images are $32\times 32$ pixels with 3 channels. During training, we apply data augmentation by padding the images with 4 zeroes on each side, then we take a random crop of $32\times 32$ and randomly flip them horizontally. At training and testing time, the inputs are normalized with $\mu=(0.4914, 0.4822, 0.4465)$, $\sigma=(0.247, 0.243, 0.261)$. In each model we evaluate on, the first convolution is kept in full-precision as well as the last linear classifier which is preceded by a ReLU activation. The models are trained with Adam \citep{kingma2014adam} and minimize the cross-entropy loss. We train a VGG-inspired architecture \citep{courbariaux2016bnn}, and a binarized ResNet-56 architecture \citep{he2015resnet} in which the $\sign$ function is applied on the input of a block instead of the output. It allows us to have a full-precision shortcut connection which limits the accuracy drop.

MobileNet-type architectures are designed for low-resource devices, and their binary training is quite challenging. This is why we also add  a third model,  MobileNet-v1 \citep{howard2017mobilenet}\footnote{\href{github.com/kuangliu/pytorch-cifar}{https://github.com/kuangliu/pytorch-cifar/blob/master/models/mobilenet.py}}  which still holds the arguments and even improves accuracy compared to Glorot's initialization.

For the VGG-inspired model the starting learning rate is $5\times10^{-3}$, otherwise it is $10^{-3}$. All models are trained for 150 epochs and their learning rate is divided by 10 at epochs 80, 120 and 140. We seek here to study how initialization from symmetric distributions with different variances affect the training of BNNs regarding the accuracy. As we showed earlier,  the variance of the latent real-valued weights of BNNs do no impact variance of activation and variance of  gradient. If we proceed with \cite{glorot2010init}, latest layers that often include more parameters are initialized closer to the origin as their variance is a decreasing function of their number of neurons.

Table~\ref{table:1} gives the training results for the models initialized from a uniform distribution with mean zero and different constant variances. We observed a similar result for non-uniform distributions, such as Gaussian which confirms Glorot's initialization is useless for BNNs.   MobileNet-v1 is already a small efficient network, so is more sensitive to compression and suffers from a more severe accuracy drop. This is where the effect of different  initialization variance is more visible: we see the largest  accuracy gain by adjusting the initialization variance.  We also tried to linearly increase and decrease the variance throughout the layers but the accuracy gain was marginal and we do not report them here. This confirms  different variance across layers has no effect in  training.
\begin{table}[]
\centering
\begin{tabular}{r  ccccc c}
  $\sigma^2$& G & $10^{-1}$  & $10^{-2}$  & $10^{-3}$ & $10^{-4}$\\ \hline
VGG  ~~ & 90.5 & 90.4 & \textbf{90.7} & 90.6 & 90.6\\ 
RN-56~~ & 88.5 & 86.9 & 88.5 & \textbf{88.8} & 88.1 \\ 
MN-v1 ~~& 71.7 & 68.9&  71.2  & 73.1 & \textbf{73.2} \\ 
\end{tabular}
\vspace{0.25cm}
\caption{Effect of different variance values on best Top1 accuracy, for binary quantized VGG, ResNet-56 (RN-56), and MobileNet version 1(MN-v1) compared to Glorot's initialization (G). }
\label{table:1}
\end{table}

In a second study, we show scaling by ${1\over \sqrt{K_{l-1}}}$ solves gradient explosion and is the main role of BatchNorm in BNN training.  As the scale does not impact the forward pass, it is interesting to see to what extent we can remove components of BatchNorm without causing too much accuracy loss. Computing the mean and variance estimates is burdensome especially on-edge-device training methods such as federated learning \citep{mcmahan2017federated}. We perform three experiments on  binarized ResNet-56. In the first experiment, features are centred but we replace the variance of BatchNorm with  constant scaling factor ${K_{l-1}}$ and we adjust it by another version (${3K_{l-1}}$) to match BatchNorm  accuracy.  In  the second experiment,  we only perform mean subtraction. In the third experiment, we replace BatchNorm layers with identity layers in the model, i.e. no BatchNorm in the network. We expect the second and the third experiment to suffer from exploding gradient and the first experiment to elevate the accuracy in a range similar to networks with BatchNorm, see Table~\ref{table:2}. 
\begin{table}[]

\centering
\begin{tabular}{l c  c c ccc}
 & BN  & ${1\over\sqrt{3K_{l-1}}}$ &${1\over\sqrt{K_{l-1}}}$   & No Scale  &  No BN \\ \hline
 & 88.8  & $87.9$& 79.6 & 31.7 & 31.7  \\ 
\end{tabular}
\vspace{0.25cm}
\caption{Ablation study of BatchNorm components on ResNet-56, maximum accuracy achieved after training using BatchNorm (BN), replacing BatchNorm with with centering and the theoretical scale  ${1\over\sqrt{K_{l-1}}}$, no scaling but only centering (No Scale), and no batch norm (No BN). The random initialization variance in each layer changes from $\sigma^2=10^{-1}$ to $10^{-4}$ as in Table~\ref{table:1} and the best results are reported.  }
\label{table:2}
\end{table}

As expected, ``No BN'' and ``No Scale'' do not converge, reaching $31.7\%$. However the accuracy reaches to  $88.8\%$ if the network has BatchNorm layers or $79.6\%$ if properly scaled, confirming our analytical study. A network with BatchNorm is still ahead in terms of accuracy, as our assumptions of initialization regarding the distributions of weights and activations become inaccurate as  training goes on.
We may close the accuracy gap by tuning the scale, for instance $1\over \sqrt{3K_{l-1}}$. BatchNorm requires floating point operations, which are costly on low-resource devices, especially while training. We recommend to replace BatchNorm with a fixed scaling factor in the order of $1\over \sqrt{K_{l-1}}$ for such devices.
\section{Conclusion}
BNNs have a different behaviour compared to their full-precision counterpart and BatchNorm has a different role in binary networks. Binary network training requires BatchNorm to converge and we clarified why it is necessary: \emph{it prevents gradient explosion}. As a biproduct of this study we found out full-precision initialization scheme is useless for BNNs. We can replace BatchNorm with a proper scaling, and we call for similar studies on more flexible fixed point quantization schemes, such as ternary, 4-bit, and 8-bit networks. 
\bibliography{references}

\begin{thebibliography}{}

\bibitem[\protect\astroncite{Alizadeh et~al.}{2019}]{alizadeh2018empiricalbnn}
Alizadeh, M., J.~Fernández-Marqués, N.~D. Lane, and
  Y.~Gal\leavevmode\nopagebreak\newline 2019.
\newblock A systematic study of binary neural networks' optimisation.
\newblock In {\em International Conference on Learning Representations}.

\bibitem[\protect\astroncite{Ardakani
  et~al.}{2019}]{ardakani2018learningrecbinter}
Ardakani, A., Z.~Ji, S.~C. Smithson, B.~H. Meyer, and W.~J.
  Gross\leavevmode\nopagebreak\newline 2019.
\newblock Learning recurrent binary/ternary weights.
\newblock In {\em International Conference on Learning Representations}.

\bibitem[\protect\astroncite{Glorot and Bengio}{2010}]{glorot2010init}
Glorot, X. and Y.~Bengio\leavevmode\nopagebreak\newline 2010.
\newblock Understanding the difficulty of training deep feedforward neural
  networks.
\newblock In {\em Proceedings of the thirteenth international conference on
  artificial intelligence and statistics}, Pp.~ 249--256.

\bibitem[\protect\astroncite{He et~al.}{2015}]{he2015resnet}
He, K., X.~Zhang, S.~Ren, and J.~Sun\leavevmode\nopagebreak\newline 2015.
\newblock Deep residual learning for image recognition.
\newblock {\em CoRR}, abs/1512.03385.

\bibitem[\protect\astroncite{Howard et~al.}{2017}]{howard2017mobilenet}
Howard, A.~G., M.~Zhu, B.~Chen, D.~Kalenichenko, W.~Wang, T.~Weyand,
  M.~Andreetto, and H.~Adam\leavevmode\nopagebreak\newline 2017.
\newblock Mobilenets: Efficient convolutional neural networks for mobile vision
  applications.
\newblock {\em CoRR}, abs/1704.04861.

\bibitem[\protect\astroncite{Hubara et~al.}{2018}]{courbariaux2016bnn}
Hubara, I., M.~Courbariaux, D.~Soudry, R.~El-Yaniv, and
  Y.~Bengio\leavevmode\nopagebreak\newline 2018.
\newblock Quantized neural networks: Training neural networks with low
  precision weights and activations.
\newblock {\em Journal of Machine Learning Research}, 18(187):1--30.

\bibitem[\protect\astroncite{Ioffe and Szegedy}{2015}]{ioffe2015batchnorm}
Ioffe, S. and C.~Szegedy\leavevmode\nopagebreak\newline 2015.
\newblock Batch normalization: Accelerating deep network training by reducing
  internal covariate shift.
\newblock {\em CoRR}, abs/1502.03167.

\bibitem[\protect\astroncite{Kingma and Ba}{2014}]{kingma2014adam}
Kingma, D.~P. and J.~Ba\leavevmode\nopagebreak\newline 2014.
\newblock Adam: A method for stochastic optimization.
\newblock {\em CoRR}, abs/1412.6980.

\bibitem[\protect\astroncite{Krizhevsky}{2009}]{krizhevsky2009cifar10}
Krizhevsky, A.\leavevmode\nopagebreak\newline 2009.
\newblock Learning multiple layers of features from tiny images.

\bibitem[\protect\astroncite{McMahan et~al.}{2017}]{mcmahan2017federated}
McMahan, H.~B., E.~Moore, D.~Ramage, S.~Hampson, and B.~A.
  y~Arcas\leavevmode\nopagebreak\newline 2017.
\newblock Communication-efficient learning of deep networks from decentralized
  data.
\newblock In {\em Proceedings of the 20th International Conference on
  Artificial Intelligence and Statistics (AISTATS)}.

\bibitem[\protect\astroncite{Santurkar et~al.}{2018}]{santurkar2018bnoptim}
Santurkar, S., D.~Tsipras, A.~Ilyas, and
  A.~Madry\leavevmode\nopagebreak\newline 2018.
\newblock How does batch normalization help optimization?
\newblock In {\em Advances in Neural Information Processing Systems 31},
  S.~Bengio, H.~Wallach, H.~Larochelle, K.~Grauman, N.~Cesa-Bianchi, and
  R.~Garnett, eds., Pp.~ 2483--2493.
\newblock Curran Associates, Inc.

\end{thebibliography}
\bibliographystyle{humannat}
\end{document}